\documentclass[11pt]{article}

\usepackage[preprint]{acl}

\usepackage{times}
\usepackage{latexsym}
\usepackage[T1]{fontenc}
\usepackage[utf8]{inputenc}
\usepackage{microtype}
\usepackage{inconsolata}

\usepackage{amsmath}
\usepackage{amssymb}
\usepackage{amsthm}
\usepackage{mathtools}
\usepackage{dsfont}

\usepackage[table]{xcolor}
\definecolor{darkred}{RGB}{170,30,30}
\definecolor{stageMI}{RGB}{56,108,176}     
\definecolor{stageTEM}{RGB}{217,95,2}      
\definecolor{stageAEB}{RGB}{27,158,119}    
\definecolor{stageSEL}{RGB}{117,112,179}   

\usepackage{booktabs}
\usepackage{multirow}
\usepackage{siunitx}
\usepackage{colortbl}

\usepackage{algorithm}
\usepackage{algpseudocode}

\usepackage{enumitem}
\usepackage{pifont}

\usepackage{natbib}

\usepackage{graphicx}

\usepackage{hyperref}

\newcommand{\methodname}{STaR-KV}

\newcommand{\stage}[2]{%
  \Statex{\color{#1}\rule[-0.2ex]{2pt}{1.6ex}}\ \textit{\color{#1}// #2}%
}

\title{STaR-KV: Spatio-Temporal Adaptive Re-weighting for KV Cache Compression in GUI Vision-Language Models}

\author{
\begin{tabular}{c@{\hspace{1.2em}}c@{\hspace{1.2em}}c@{\hspace{1.2em}}c}
\textbf{Yuhang Han}\textsuperscript{1,2*} &
\textbf{Wenzheng Yang}\textsuperscript{3*} &
\textbf{Yujie Chen}\textsuperscript{1,2} &
\textbf{Xiangqi Jin}\textsuperscript{1,2} \\
\multicolumn{4}{c}{
\textbf{Yaojie Zhang}\textsuperscript{1,4} \quad
\textbf{Siteng Huang}\textsuperscript{5} \quad
\textbf{Linfeng Zhang}\textsuperscript{1\textdagger}
}
\end{tabular}
\\[0.5em]
\textsuperscript{1}EPIC Lab, SJTU \quad
\textsuperscript{2}HKUST (GZ) 
\\
\textsuperscript{3}The University of Sydney \quad
\textsuperscript{4}UESTC \quad
\textsuperscript{5}ZJU
\\[0.3em]
\textsuperscript{*}Equal contribution. \quad
\textsuperscript{\textdagger}Corresponding author.
}

\begin{document}
\maketitle
\begin{abstract}
Vision-language-model-based graphical user interface (GUI) agents
have shown broad automation capabilities, yet deployment is
bottlenecked by a key-value (KV) cache that grows linearly with
interaction steps. For instance, UI-TARS-1.5-7B consumes $76$\,GB of GPU memory on
merely five screenshots, approaching the capacity of mainstream
80\,GB accelerators. Existing KV
compression methods share two structural assumptions: aggregating
visual-token importance into a \emph{single shared saliency map},
and applying a \emph{fixed top-$B$ cutoff} to the fused score
distribution. Pilot measurements refute both: spatial specialization
lives at the attention-subspace level and migrates across layers,
while the score distribution drifts in shape along a trajectory. We
propose \textbf{STaR-KV} (\textbf{S}patio-\textbf{T}emporal
\textbf{A}daptive \textbf{R}e-weighting), a training-free KV cache
compression framework that calibrates token importance along three
axes: (i) subspace-aware scoring driven by online spatial mutual
information; (ii) a temporal stability discount that suppresses
redundant cache entries from persistently attended subspaces; and
(iii) an entropy-derived temperature that adaptively reshapes the
score distribution. Across four GUI benchmarks, STaR-KV achieves the strongest average
accuracy among state-of-the-art KV compression methods (e.g.,
GUIKV, SnapKV) at matched budgets, with no compression-stage
FLOPs overhead ($-0.07\%$) and cutting peak GPU memory by nearly
$40\%$ at a $20\%$ KV-cache budget. Code is available at \url{https://github.com/kawhiiiileo/STaR-KV}.
\end{abstract}


\section{Introduction}
\label{sec:introduction}
Graphical user interface (GUI) agents are vision-language model
(VLM)~\cite{innovatorvl,qwen,qwen3vl} based systems that automate
human-computer workflows by observing screenshots and predicting
interaction actions, and have demonstrated remarkable capabilities
across desktop, mobile, and web environments~\cite{uitars,opencua,seeclick,cogagent}.
Deploying such systems, however, remains prohibitively expensive: a
single task typically ingests tens of high-resolution screenshots as
historical context, causing the KV cache to grow linearly with the
number of interaction steps. For instance, UI-TARS-1.5-7B consumes $76$\,GB of GPU memory on
merely five screenshots~\cite{GUIKV}, nearly exhausting a single
80\,GB GPU.

A natural and training-free remedy is KV cache compression. Prior work
has progressed through three increasingly specialized stages. Early
efforts on \emph{general LLM KV compression}~\cite{streamingLLM,h2o,snapkv,pyramidkv}
 score tokens by attention statistics but treat all modalities uniformly. \emph{General
VLM KV compression}~\cite{lookm,vlcache,MEDA} adds
modality-aware budgeting and KV merging, yet targets natural images
and videos rather than GUIs. Most recently, \emph{GUI-aware KV
compression}~\cite{GUIKV} introduces a two-axis design
combining residual-stream saliency with pairwise frame redundancy,
achieving strong accuracy without retraining.

\begin{figure}[t]
  \centering
  \includegraphics[width=\columnwidth]{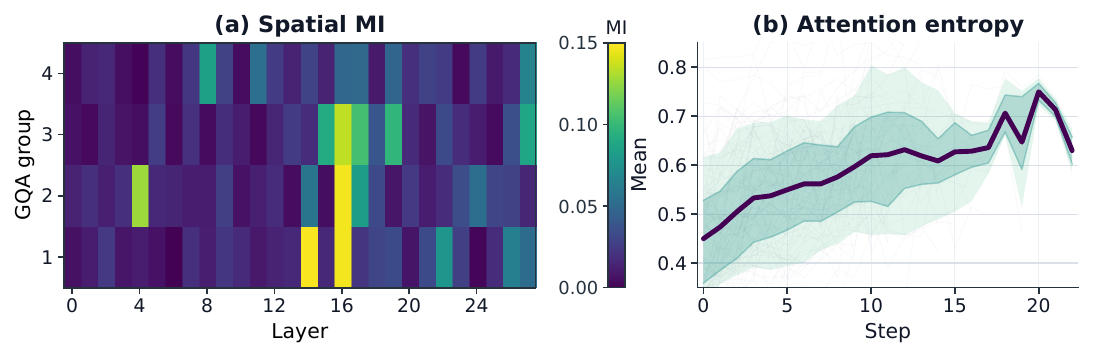}
  \caption{Pilot measurements on UI-TARS-1.5-7B.
\textbf{(a)} Per-group MI with 2D screen coordinates;
red arrows mark spatially dominant GQA groups migrating across layers.
\textbf{(b)} Normalized attention entropy $\hat{H}$ over trajectory
steps; dark line: mean, shaded band: inter-trajectory variance, thin
traces: individual trajectories.}
  \label{fig:introduction}
\end{figure}

Despite this progress, all existing methods share two structural assumptions that become problematic for GUI reasoning. First, they aggregate visual-token importance into a single shared saliency map, treating all attention heads (or GQA groups) as spatially homogeneous. Second, they apply a fixed top-B cutoff to the fused score distribution, assuming a stable shape across frames. We argue that both assumptions are systematically violated in GUI agents, as evidenced by the following pilot measurements on UI-TARS-1.5-7B~\cite{uitars} with AgentNetBench trajectories (Fig~\ref{fig:introduction}).



\noindent\textit{\textbf{Pilot 1: Subspace-level spatial specialization.}}
We measure the mutual information (MI) between per-subspace
attention patterns and 2D screen coordinates; a high MI indicates
that a subspace's attention is spatially localized rather than
uniformly spread, i.e., it encodes layout-sensitive structure.
Figure~\ref{fig:introduction}(a) shows that, within the same layer,
the most spatially coupled subspace (a GQA group~\cite{gqa} here;
per-head results in Fig.~\ref{fig:gqavsmha}) attains
$3$–$7\times$ higher MI than the weakest, with the dominant
subspace migrating across layers. The ranking is stable
across frames (Spearman $\rho>0.85$), confirming a signal that
shared saliency erases and that is estimable online.

\noindent\textit{\textbf{Pilot 2: Entropy drift and inter-trajectory dispersion.}}
The normalized attention entropy $\hat{H}\in[0,1]$
(Figure~\ref{fig:introduction}(b)) rises monotonically and plateaus around
step~$10$, indicating progressive flattening; meanwhile the variance
band spans $0.2$–$0.3$ at every step, indicating that two tasks at
the same step can sit in markedly different regimes. A fixed top-$B$
must therefore adapt both across tasks and along steps, and is
structurally unreliable.

Together, Pilot 1 reveals a \emph{spatial} blind spot (subspace-level
heterogeneity is erased) and Pilot 2 reveals a \emph{distributional}
blind spot (fixed cutoffs fail under shape drift). A natural
corollary of Pilot 1 is that tokens governed by temporally stable
subspaces (e.g., persistent toolbars) accumulate redundant cache
entries across frames, exposing a third, \emph{temporal} axis that
single-frame spatial profiling alone cannot capture.

Motivated by these findings, we propose \textbf{STaR-KV}
(\textbf{S}patio-\textbf{T}emporal \textbf{A}daptive
\textbf{R}e-weighting), a training-free KV cache compression
framework that addresses the three blind spots along three
complementary calibration axes. Our contributions are summarized
below.

\noindent\ding{172}\ \textbf{Online Spatial Profiling.}
We introduce subspace-aware scoring driven by an online estimate of spatial mutual information between attention and 2D screen coordinates, preserving layout-sensitive signals at the subspace level.

\noindent\ding{173}\ \textbf{Cumulative Temporal Stability Discount.}
We propose a temporal discount that weights historical visual tokens by the cumulative stability of their governing subspace across frames, suppressing stale cache entries from persistent UI structures while preserving tokens attached to dynamic widgets.

\noindent\ding{174}\ \textbf{Adaptive Entropy-Based Sharpening (AEB).}
We reshape the score distribution through a temperature derived from
its normalized entropy, eliminating the arbitrariness of fixed
top-$B$ cutoffs under flat regimes while sharpening selection under
peaky ones, at no extra memory budget.

Across diverse GUI agent benchmarks, STaR-KV requires no
fine-tuning and is directly applicable to existing GUI VLMs such as
UI-TARS and OpenCUA, achieving stronger average performance than
state-of-the-art KV compression methods (e.g., GUIKV, SnapKV) under
matched memory budgets.

\section{Related Work}

\subsection{KV Cache Compression}

KV cache compression reduces inference memory by deciding which historical
key--value states to retain. In LLMs, methods such as StreamingLLM, H2O,
SnapKV, PyramidKV, and FastGen make this decision without retraining, using
attention sinks, heavy hitters, recent-window attention, layer-wise budgets, or
head-specific policies~\citep{streamingLLM,h2o,snapkv,pyramidkv,fastgen,kawhi}. A parallel line uses cross-step stability as a redundancy signal to accelerate long-context prefilling~\citep{dash}. 
VLM-oriented methods~\cite{ficoco,globalcom} bring the same idea to multimodal inputs, where visual and
text tokens differ in sparsity and redundancy, through modality-aware scoring,
dynamic layer allocation, or KV merging~\citep{lookm,vlcache,MEDA}. These
signals are useful for GUI agents as well, but GUI histories add a more regular
structure: the same layout may persist for many steps, while small widgets or
action targets matter only in particular task states. STaR-KV targets this
setting by calibrating cache scores with subspace-aware spatial profiles,
cumulative temporal stability, and entropy-aware selection.

\subsection{Efficient GUI Agents}

Complementary GUI-agent efficiency methods act on the input or context rather
than on the accumulated KV cache. ShowUI selects UI-aware visual tokens,
SimpAgent simplifies dense element and history context, and GUIPruner prunes
high-resolution screenshot streams with temporal and structural
cues~\citep{showui,simpagent,guipruner}. These methods exploit layout,
saliency, and recency to shorten or sparsify the screenshot/history stream;
they do not directly decide which cached key--value states to retain.

Closer to our setting, GUIKV~\citep{GUIKV} compresses GUI caches with spatial
saliency and temporal redundancy pruning, while ST-Lite uses component-centric
saliency and trajectory-aware semantic gating~\citep{stlite}. Both suggest
that GUI-specific spatial--temporal cues are stronger than generic KV signals.
STaR-KV follows this cache-side direction, but changes the granularity of the
decision: it refines cache ranking through per-head or GQA-group spatial
profiling, cumulative group-level temporal stability, and entropy calibration
before fixed-budget selection.

\section{Method}
\begin{figure*}[t]
  \centering
  \includegraphics[width=\textwidth]{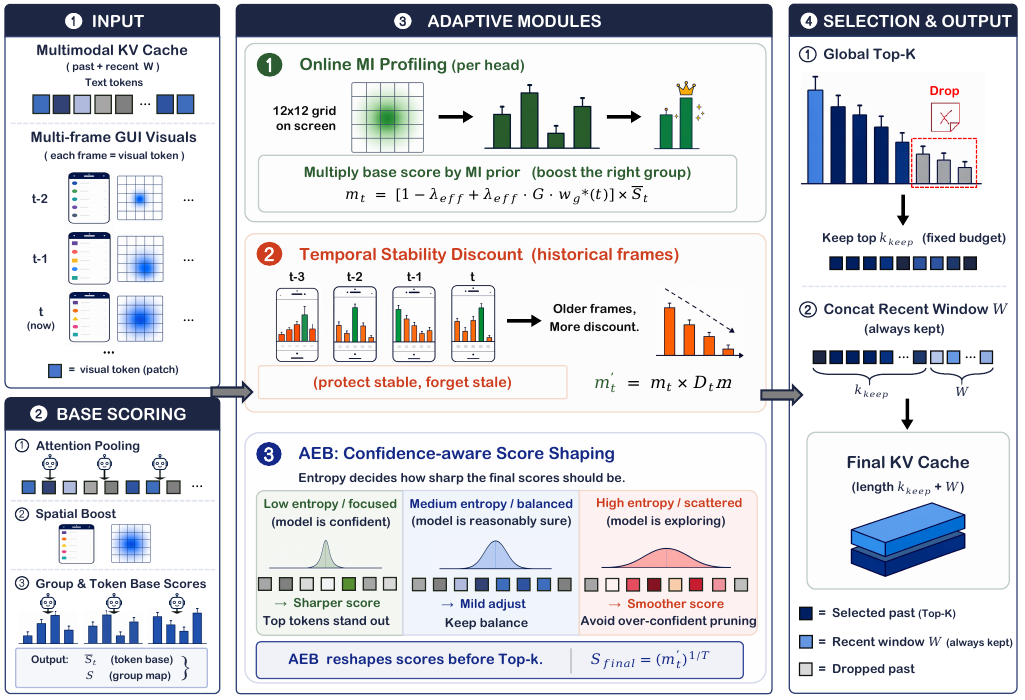}
  \caption{\textbf{Overview of \methodname{}.} Given a multimodal KV cache of
    text tokens and multi-frame GUI visual tokens (\textbf{\ding{182} Input}),
    \methodname{} (i) builds a token-level \emph{base score} $\bar{s}_t$ and a
    GQA group score map $\mathbf{S}$ from pooled recent-query attention with
    an optional GUI spatial boost (\textbf{\ding{183} Base Scoring}), and then
    (ii) refines this base score along three complementary axes
    (\textbf{\ding{184} Adaptive Modules}): \textbf{Online MI Profiling}
    estimates a per-group 2D mutual-information prior $w_{g^{*}(t)}$ and uses
    it to up-weight the group that best explains each token; the
    \textbf{Temporal Stability Discount} multiplies the score by a cumulative
    decay $D_t$ that discounts stale tokens governed by stable groups while
    preserving tokens tied to dynamic groups; and \textbf{Adaptive Entropy-Based
    Sharpening} (AEB) reshapes the fused score by an entropy-derived
    temperature $1/T$---sharpening when attention is focused and smoothing
    when it is scattered---without changing the cache budget. Finally
    (\textbf{\ding{185} Selection \& Output}), a single global Top-$K$ over
    the final score $s_{\mathrm{final}}$ retains $k_{\text{keep}}$ historical
    tokens, which are concatenated with the always-kept recent window $W$ to
    form the compressed KV cache.}
  \label{fig:method_overview}
\end{figure*}
 
\subsection{Preliminaries}
\label{sec:prelim}

\paragraph{GUI token stream.}
At each step, a GUI agent conditions on screenshots, the user instruction, and
previous actions, then autoregressively predicts the next action. We index the
historical context by token position $t\in\{1,\ldots,L\}$. Each token is either
non-visual (text or action history) or visual. A visual token inherits a frame
index $f(t)$ and a discretized screen coordinate $(r_t,c_t)$; non-visual tokens
have no screen coordinate. Let $\mathds{1}_{\text{vis}}(t)$ indicate whether
token $t$ is visual. For current frame $F$, the frame distance of a visual token
is $\Delta_{\text{f}}(t)=F-f(t)$.

\paragraph{KV cache and attention subspaces.}
Once the historical context has been processed, each transformer layer stores
the corresponding key and value projections in the KV cache. With grouped-query
attention (GQA)~\citep{gqa}, $H$ query heads share $G$ key--value groups
($G<H$), so
$\mathbf{K},\mathbf{V}\in\mathbb{R}^{G\times L\times d}$. Query head $h$
attends to the KV group
$g(h)=1+\lfloor (h-1)G/H\rfloor$:
\begin{equation}
\mathbf{A}[h,:]
  = \mathrm{softmax}\!\Bigl(
      \tfrac{\mathbf{Q}_h\,\mathbf{K}_{g(h)}^{\top}}{\sqrt{d}}
    \Bigr),
\quad
\mathbf{A}\in\mathbb{R}^{H\times L}.
\end{equation}
Standard multi-head attention is recovered by setting $G=H$ and $g(h)=h$. We
call the finest cache-specific attention unit exposed by the architecture a
\emph{subspace}: an individual head for MHA, or a shared KV group for GQA.
Different subspaces may specialize in layout, text semantics, or
action-relevant history.

\paragraph{Compression interface.}
We use the standard fixed-budget interface for training-free KV compression.
Given cache length $L$ and a historical-token budget $B\!\ll\!L$, the method
assigns each historical token an importance score $s_t$ and keeps the top-$B$
scored tokens. A recent window of size $W_{\text{rec}}$ is retained outside this
budget to preserve the immediate decoding context. Existing attention-based
compressors usually compute $s_t$ from recent-query attention over the cache.
STaR-KV keeps the same interface, but calibrates the score with subspace, frame,
and distribution-level signals tailored to GUI trajectories.

\subsection{STaR-KV}
\label{sec:overview}

We consider a vision-language model that maintains a KV cache of
length $L$ over a stream of GUI frames and must, under a fixed
budget $B\!\ll\!L$, retain the most informative historical
key--value pairs. Motivated by the three blind spots identified in
Sec.~\ref{sec:introduction}, we propose \textbf{STaR-KV}
(Fig~\ref{fig:method_overview}), a training-free \textbf{three-axis}
pipeline that refines a base attention score $\bar{s}_t$
(\S\ref{sec:base_score}) along three complementary axes:
\emph{spatial} (a per-subspace mutual-information prior;
\S\ref{sec:mi_profiling}), \emph{temporal} (a cumulative stability
discount; \S\ref{sec:temporal}), and \emph{distributional} (an
adaptive entropy-based sharpening, AEB; \S\ref{sec:aeb}). Letting
$\mathds{1}_{\text{vis}}(t)$ indicate that token $t$ is a historical
visual token, the final score takes the form
\begin{equation}
\label{eq:final_score}
\begin{aligned}
s_t^{(\mathrm{f})}
   &= \bar{s}_t \cdot \beta_t \cdot D_t^{\,\mathds{1}_{\text{vis}}(t)},\\[2pt]
\beta_t
   &= (1{-}\lambda^{(t)}) + \lambda^{(t)}\, G\, w_{g^{*}(t)},\\[2pt]
s_t &= \max\bigl(s_t^{(\mathrm{f})},\, 0\bigr)^{1/T}.
\end{aligned}
\end{equation}
The top-$B$ tokens under $s_t$, concatenated with a fixed window of
recent tokens, form the compressed cache (\S\ref{sec:selection}). All
statistics are maintained online, per layer, and incur only $O(L)$
additional memory. We next describe the base score construction
(\S\ref{sec:base_score}), followed by the three calibration axes
(\S\ref{sec:mi_profiling}--\ref{sec:aeb}) and the unified selection
step (\S\ref{sec:selection}).


\subsection{Base Score Construction}
\label{sec:base_score}

Given a KV cache of length $L$, we aggregate attention from the $W$
most recent queries against all historical keys and apply a $1$D
average pool to suppress positional jitter:
\begin{align}
\mathbf{A}_{\text{raw}}
  &= \sum_{i=0}^{W-1}
     \mathrm{softmax}\!\Bigl(
        \tfrac{\mathbf{Q}_{-i}\mathbf{K}_{\text{past}}^{\top}}{\sqrt{d}}
     \Bigr),\\
\mathbf{A}
  &= \mathrm{AvgPool1D}(\mathbf{A}_{\text{raw}})
   \in \mathbb{R}^{H\times L}.
\end{align}
We then collapse the $H$ heads into $G$ GQA groups and average within
each group, obtaining $\mathbf{S}\in\mathbb{R}^{G\times L}$, and finally
average across groups to yield the token-level base score:
\begin{equation}
\mathbf{S}[g,t]
  = \tfrac{G}{H}\!\!\sum_{h=(g-1)H/G}^{gH/G-1}\!\!\mathbf{A}[h,t],
\quad
\bar{s}_t = \tfrac{1}{G}\sum_{g=1}^{G}\mathbf{S}[g,t].
\label{eq:base_score}
\end{equation}
Averaging over groups in Eq.~\eqref{eq:base_score} erases the per-group
spatial specialisation we wish to exploit; the next subsection recovers
it. We additionally support an \emph{optional} residual-stream
augmentation in which hidden-state norms $u_i\!=\!\|\mathbf{h}_i\|_2$
inside the visual span are added to $\mathbf{A}_{h,i}$ with weight
$\alpha$; $\alpha\!=\!0$ disables it and is the default unless stated
otherwise. We refer to each such group as a \emph{subspace} in the rest of this
paper; for MHA-based~\cite{MHA} models, each subspace coincides with an individual head.


\subsection{Online Spatial Profiling}
\label{sec:mi_profiling}

To recover the per-subspace spatial heterogeneity identified in
Sec.~\ref{sec:introduction}, we estimate each subspace's coupling
with screen coordinates via online mutual information.

\textbf{Online MI estimation.}
For each subspace $g$ (an individual head under MHA or a shared KV
group under GQA), we bin its attention values
$\{\mathbf{S}[g,t]\}_{t\in\mathrm{vis}}$ into $K$ equal-frequency
bins and pair them with the discretised row/column index
$(r_t,c_t)$ of each visual token on an $M\!\times\!M$ grid. With
$\hat{p}$ the joint histogram, the plug-in estimator and its
exponential moving average (EMA) across frames are
\begin{equation}
\mathrm{MI}_g
   = \sum_{b,r,c}\hat{p}(b,r,c)\,
     \log\frac{\hat{p}(b,r,c)}{\hat{p}(b)\,\hat{p}(r,c)},
\end{equation}
\begin{equation}
\overline{\mathrm{MI}}_g^{(t)}
   = \rho\,\overline{\mathrm{MI}}_g^{(t-1)}
   + (1-\rho)\,\mathrm{MI}_g^{(t)}.
\end{equation}
The EMA stabilises the estimator and damps the feedback loop between
attention and its own re-weighting.

\textbf{Subspace prior and re-weighting.}
With
$\Delta_g \!=\! \overline{\mathrm{MI}}_g \!-\! \min_j\overline{\mathrm{MI}}_j$
and
$\Delta_{\max} \!=\! \max_j\overline{\mathrm{MI}}_j \!-\! \min_j\overline{\mathrm{MI}}_j$,
\begin{equation}
\tilde{m}_g = \frac{\Delta_g}{\Delta_{\max}+\varepsilon},
\quad
w_g = \mathrm{softmax}(\tilde{m}_g / \tau).
\end{equation}
Each token is bound to its most responsive subspace,
$g^{*}(t) = \arg\max_g \mathbf{S}[g,t]$, giving the MI-modulated score
$s_t^{(\mathrm{mi})} = \bar{s}_t \cdot \beta_t$ with $\beta_t$ from
Eq.~\eqref{eq:final_score}. The convex combination preserves scale
($G w_{g^*}$ averages to $1$ when $w$ is uniform), and $\lambda^{(t)}$
is linearly annealed from $0$ to the target MI prior strength $\lambda$
over the first $N_{\text{warmup}}{+}N_{\text{ramp}}$ frames so that
the prior acts only after its EMA has converged.

\subsection{Cumulative Temporal Stability Discount}
\label{sec:temporal}

MI profiling is computed per frame and cannot distinguish a
persistent toolbar from a transient pop-up. We track temporal
stability \emph{cumulatively} via a per-subspace attention EMA,
letting each KV entry inherit the temporal credit of frames in
which its governing subspace remained consistent.

\textbf{Stability and decay.}
Let $\mathbf{s}_g^{(\text{c})}$ denote the current row of $\mathbf{S}$
for subspace $g$ and $\mathbf{m}_g$ its EMA-tracked historical mean.
The stability and its update are
\begin{equation}
\phi_g = \cos\bigl(\mathbf{s}_g^{(\text{c})},\,\mathbf{m}_g\bigr),
\quad
\mathbf{m}_g \leftarrow \rho_{\text{t}}\,\mathbf{m}_g
       + (1-\rho_{\text{t}})\,\mathbf{s}_g^{(\text{c})}.
\end{equation}
A high $\phi_g$ indicates a temporally consistent attention pattern
(static UI chrome); a low $\phi_g$ signals burst-like activation
(transient widgets). We apply the resulting decay only to historical
visual tokens; non-visual tokens retain $D_t\!=\!1$. With
$\Delta_{\text{f}}(t)$ the frame distance of token $t$ to the current
frame and $\delta$ the temporal decay rate,
\begin{equation}
D_t = D_{\min} + (1-D_{\min})\,e^{-\eta_t},
\quad
\eta_t = \delta\,\Delta_{\text{f}}(t)\,\phi_{g^{*}(t)}.
\end{equation}
Tokens governed by stable subspaces decay rapidly as they age, while
tokens attached to dynamic subspaces are shielded; $D_{\min}$ caps the
maximal discount. Combining with the MI step yields
$s_t^{(\text{temp})} = s_t^{(\mathrm{mi})}\!\cdot\! D_t$, equal to
$s_t^{(\mathrm{f})}$ in Eq.~\eqref{eq:final_score}.


\subsection{Adaptive Entropy-Based Sharpening}
\label{sec:aeb}

Even after the spatial (MI) and temporal (TEM) axes, the calibrated
score distribution can be too flat (making the top-$B$ cut arbitrary)
or too peaked (discarding useful mid-ranked tokens). We close this
gap with a single, data-dependent temperature derived from the
\emph{base} distribution, so that AEB acts as an independent global
calibrator. With
$Z = \sum_{t'}\bar{s}_{t'}$ and $p_t = \bar{s}_t / Z$,
\begin{equation}
\begin{split}
\hat{H} &= -\frac{1}{\log L}\sum_{t=1}^{L} p_t\log p_t,\\[2pt]
T &= T_{\min} + (T_{\max}-T_{\min})\,\hat{H},
\end{split}
\end{equation}
and the temperature is applied as a power,
$s_t = \max(s_t^{(\text{temp})},0)^{1/T}$. Concentrated base scores
($\hat{H}\!\to\!0$) yield $T\!\ll\!1$ and sharpen $s_t$; diffuse base
scores ($\hat{H}\!\to\!1$) yield $T\!>\!1$ and flatten $s_t$ to avoid
over-commitment to a narrow set of tokens. We refer to
$[T_{\min}, T_{\max}]$ as the \emph{AEB $T$ range}.

\begin{table*}[t!]
\centering
\small
\renewcommand{\arraystretch}{1.05}
\setlength{\tabcolsep}{3pt}
\caption{Main results on four GUI-agent benchmarks under four KV-cache budgets ($5\%$, $10\%$, $20\%$, $40\%$) for UI-TARS-1.5-7B and OpenCUA-7B. The top ``Full Cache'' row reports uncompressed accuracy for reference. Dataset abbreviations: \textbf{SS-Pro}=ScreenSpot-Pro, \textbf{SS-v2}=ScreenSpot-v2, \textbf{AndC}=AndroidControl, \textbf{ANB}=AgentNetBench. \textbf{Avg.} is the mean across the four datasets; relative accuracy change from full cache is shown in parentheses ($\uparrow$/$\downarrow$). STaR-KV rows are highlighted in \textcolor{cyan}{cyan}. All scores are in percentage (\%); higher is better. The best result in each (dataset, budget) cell per model is in \textbf{bold}; ties are bolded jointly.}
\label{tab:main_results}
\resizebox{\textwidth}{!}{%
\begin{tabular}{l ccccc|ccccc}
\toprule
& \multicolumn{5}{c|}{\textbf{UI-TARS-1.5-7B}} & \multicolumn{5}{c}{\textbf{OpenCUA-7B}} \\
\cmidrule(lr){2-6} \cmidrule(lr){7-11}
\textbf{Method} & \textbf{Avg.} & \textbf{SS-Pro} & \textbf{SS-v2} & \textbf{AndC} & \textbf{ANB} & \textbf{Avg.} & \textbf{SS-Pro} & \textbf{SS-v2} & \textbf{AndC} & \textbf{ANB} \\
\midrule
Full Cache & 49.75 & 41.68 & 88.79 & 48.20 & 20.34 & 51.80 & 46.81 & 91.13 & 37.80 & 31.45 \\
\midrule
\rowcolor[gray]{0.9}
\multicolumn{11}{l}{\emph{Budget = 40\%}} \\
PyramidKV               & 45.45 \scriptsize($\downarrow$8.64\%)          & 38.71          & 87.58          & 45.20          & 10.30          & 49.47 \scriptsize($\downarrow$4.50\%)          & 45.86          & 91.02          & 35.80          & 25.18 \\
SnapKV                  & 47.22 \scriptsize($\downarrow$5.09\%)          & 41.11          & 88.34          & 42.00          & 17.44          & 49.32 \scriptsize($\downarrow$4.79\%)          & 44.28          & 91.02          & \textbf{36.80} & 25.18 \\
GUIKV                   & 48.92 \scriptsize($\downarrow$1.67\%)          & 40.73          & 88.51          & 46.40          & 20.02          & 47.31 \scriptsize($\downarrow$8.67\%)          & 44.28          & 89.10          & 35.20          & 20.67 \\
\rowcolor{cyan!10}
\textbf{STaR-KV (Ours)} & \textbf{49.94} \scriptsize(\textbf{$\uparrow$0.38\%}) & \textbf{41.70} & \textbf{88.95} & \textbf{48.00} & \textbf{21.10} & \textbf{50.04} \scriptsize(\textbf{$\downarrow$3.40\%}) & \textbf{46.68} & \textbf{91.05} & \textbf{36.80} & \textbf{25.64} \\
\midrule
\rowcolor[gray]{0.9}
\multicolumn{11}{l}{\emph{Budget = 20\%}} \\
PyramidKV               & 39.46 \scriptsize($\downarrow$20.68\%)         & 35.36          & 84.74          & 34.20          & 3.52           & 45.30 \scriptsize($\downarrow$12.55\%)         & 42.57          & 88.80          & 31.80          & 18.04 \\
SnapKV                  & 45.09 \scriptsize($\downarrow$9.37\%)          & 39.22          & 88.28          & 41.80          & 11.04          & 41.17 \scriptsize($\downarrow$20.52\%)         & 45.48          & 67.96          & 33.20          & 18.04 \\
GUIKV                   & 46.59 \scriptsize($\downarrow$6.35\%)          & 39.85          & 88.46          & \textbf{43.00} & 15.06          & 43.03 \scriptsize($\downarrow$16.93\%)         & 40.04          & 88.04          & 28.80          & 15.23 \\
\rowcolor{cyan!10}
\textbf{STaR-KV (Ours)} & \textbf{47.31} \scriptsize(\textbf{$\downarrow$4.90\%}) & \textbf{40.90} & \textbf{88.90} & \textbf{43.00} & \textbf{16.45} & \textbf{47.01} \scriptsize(\textbf{$\downarrow$9.25\%}) & \textbf{45.60} & \textbf{89.92} & \textbf{34.40} & \textbf{18.10} \\
\midrule
\rowcolor[gray]{0.9}
\multicolumn{11}{l}{\emph{Budget = 10\%}} \\
PyramidKV               & 33.12 \scriptsize($\downarrow$33.43\%)         & 29.48          & 75.77          & 25.40          & 1.84           & 39.65 \scriptsize($\downarrow$23.46\%)         & 38.96          & 86.29          & 24.20          & 9.14 \\
SnapKV                  & 40.04 \scriptsize($\downarrow$19.52\%)         & 35.74          & 85.32          & 35.00          & 4.11           & 38.27 \scriptsize($\downarrow$26.12\%)         & 44.09          & 72.11          & \textbf{29.60} & 7.29 \\
GUIKV                   & 40.70 \scriptsize($\downarrow$18.19\%)         & 36.05          & 85.35          & \textbf{36.40} & \textbf{5.00}  & 39.16 \scriptsize($\downarrow$24.40\%)         & 37.82          & 86.02          & 23.40          & 9.39 \\
\rowcolor{cyan!10}
\textbf{STaR-KV (Ours)} & \textbf{40.72} \scriptsize(\textbf{$\downarrow$18.15\%}) & \textbf{37.70} & \textbf{85.39} & 35.20          & 4.60           & \textbf{42.67} \scriptsize(\textbf{$\downarrow$17.63\%}) & \textbf{44.21} & \textbf{87.01} & 29.20          & \textbf{10.26} \\
\midrule
\rowcolor[gray]{0.9}
\multicolumn{11}{l}{\emph{Budget = 5\%}} \\
PyramidKV               & 24.70 \scriptsize($\downarrow$50.35\%)         & 20.75          & 57.39          & 19.80          & 0.84           & 32.72 \scriptsize($\downarrow$36.83\%)         & 33.90          & 81.16          & 13.00          & 2.81 \\
SnapKV                  & 32.04 \scriptsize($\downarrow$35.60\%)         & 29.22          & 76.71          & 21.00          & 1.23           & 33.62 \scriptsize($\downarrow$35.14\%)         & 38.77          & 73.11          & \textbf{19.80} & 2.81 \\
GUIKV                   & 32.63 \scriptsize($\downarrow$34.41\%)         & 30.30          & 77.15          & \textbf{21.80} & 1.25           & 33.78 \scriptsize($\downarrow$34.79\%)         & 32.95          & 81.84          & 17.40          & 2.92 \\
\rowcolor{cyan!10}
\textbf{STaR-KV (Ours)} & \textbf{32.74} \scriptsize(\textbf{$\downarrow$34.19\%}) & \textbf{30.50} & \textbf{77.54} & 21.60          & \textbf{1.30}  & \textbf{35.89} \scriptsize(\textbf{$\downarrow$30.71\%}) & \textbf{38.96} & \textbf{83.58} & 17.60          & \textbf{3.42} \\
\bottomrule
\end{tabular}}
\vspace{-2mm}
\end{table*}

\subsection{Unified Selection}
\label{sec:selection}

We retain the top-$B$ historical tokens under $s_t$ and concatenate
them with a fixed window of $W_{\text{rec}}$ recent tokens (kept
unconditionally and counted outside $B$):
\begin{equation}
\mathcal{I} = \mathrm{TopK}\bigl(\{s_t\}_{t=1}^{L},\,B\bigr),
\end{equation}
\begin{equation}
\mathbf{K}'
  = \bigl[\mathbf{K}_{\text{past}}[\mathcal{I}];\,
          \mathbf{K}_{\text{rec}}\bigr],\;
\mathbf{V}'
  = \bigl[\mathbf{V}_{\text{past}}[\mathcal{I}];\,
          \mathbf{V}_{\text{rec}}\bigr].
\end{equation}
All modules are training-free, operate per layer, and add no learnable
parameters; their overhead is dominated by the $O(L)$ EMA buffers and
the $G\!\times\!K\!\times\!M^2$ joint histogram, both negligible
compared to the cache itself.
\section{Experiment}

\subsection{Experimental Setup}

\paragraph{Models.}
We instantiate STaR-KV on two open-source GUI agents:
UI-TARS-1.5-7B~\cite{uitars} for general GUI grounding, and
OpenCUA-7B~\cite{opencua} for longer-horizon computer-use tasks,
covering both short grounding queries and complex trajectories. We
use the official checkpoints, prompts, and action formats, and apply
STaR-KV only at inference.

\paragraph{Benchmarks.}
Our evaluation spans four GUI-agent benchmarks covering single-step
grounding and trajectory-based action prediction.
ScreenSpot-Pro~\cite{screenspotpro} and
ScreenSpot-v2~\cite{screenspotpro} assess grounding from a
screenshot and an instruction, targeting professional desktop and
mobile/desktop/web scenarios respectively. AndroidControl~\cite{androidcontrol}
and AgentNetBench~\cite{opencua} test whether compressed caches
preserve decision-relevant history over recorded trajectories.

\paragraph{Cache budgets and metrics.}
All methods are compared at four cache budgets $\{5,10,20,40\}\%$.
Each benchmark uses its canonical metric: \emph{grounding accuracy}
on ScreenSpot-Pro and ScreenSpot-v2 (predicted click inside the
ground-truth box), \emph{step accuracy} on AndroidControl (matching
action type and normalized arguments), and the official
\emph{per-step action score} on AgentNetBench.

\begin{figure*}[t]
  \centering
  \includegraphics[width=\textwidth]{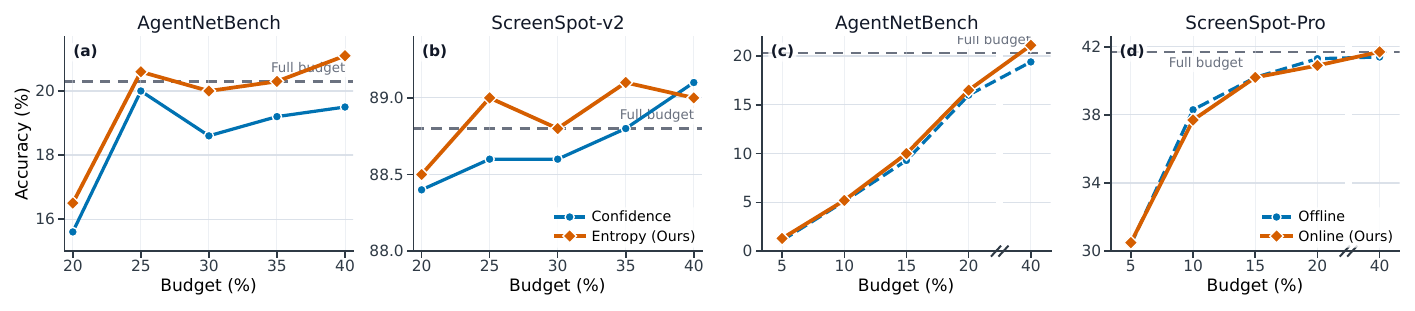   }
  \caption{\textbf{Ablation studies of \methodname{} on  UI-TARS-1.5-7B~\cite{uitars} under different
    KV cache budgets.}
    (a)--(b): entropy-based vs.\ confidence-based AEB on AgentNetBench and
    ScreenSpot-v2; (c)--(d): online (ours) vs.\ offline group prior on
    AgentNetBench~\cite{opencua} and ScreenSpot-Pro~\cite{screenspotpro}. The horizontal axis is the cache budget
    (\%) and the vertical axis is the corresponding accuracy (\%). Our
    entropy-based AEB and online group prior consistently match or outperform
    their counterparts across budgets, and the gains are most pronounced on
    long-horizon agent trajectories (AgentNetBench).}
  \label{fig:ablation_4panel}
  \vspace{-2mm}
\end{figure*}

\subsection{Experimental Results}

\noindent\textbf{Main results.}
As shown in Table~\ref{tab:main_results}, STaR-KV achieves the
best average accuracy across datasets under most compression levels.
For UI-TARS-1.5-7B~\cite{uitars}, STaR-KV's average accuracy ranges
from $49.94$ to $32.74$, yielding the strongest average result among
PyramidKV~\cite{pyramidkv}, SnapKV~\cite{snapkv}, and
GUIKV~\cite{GUIKV} at every listed budget; notably, its average score
surpasses even the full-cache baseline ($49.75$) at the $40\%$ budget
(\textcolor{red}{$\uparrow$1.02} over GUIKV), benefiting from
fine-grained spatial priors that filter redundant visual noise. For OpenCUA-7B~\cite{opencua}, STaR-KV
exhibits the \textbf{smallest degradation} across all budgets,
leading GUIKV by \textcolor{red}{$\uparrow$2.1}–\textcolor{red}{$\uparrow$4.0}
percentage points at low budgets (e.g., \textcolor{red}{$\uparrow$3.98}
at $20\%$). This advantage is especially pronounced in action-oriented
settings, where dense visual inputs and fluctuating attention make
cache ranking harder, precisely the regime that STaR-KV's spatial
priors and entropy shaping are designed to address. Overall, STaR-KV
exhibits the most graceful accuracy decay, validating the joint
calibration value of fine-grained spatial priors, temporal
discounting, and entropy-guided shaping before global Top-$K$
eviction.
\begin{table}[t]
  \centering
  \small
  \setlength{\tabcolsep}{4pt}
  \renewcommand{\arraystretch}{0.85}
  \caption{Ablation on the three calibration modules of STaR-KV. \textbf{Base}
denotes the GQA-averaged attention score $\bar{s}_t$ with global
Top-$K$ selection, prior to any calibration. Cell values are step
accuracy (\%); column headers indicate the KV-cache budget.}
  \label{tab:main_ablation}
  \resizebox{\columnwidth}{!}{
  \begin{tabular}{@{}l cc cc@{}}
  \toprule
  \multirow{2}{*}{\textbf{Method}}
    & \multicolumn{2}{c}{\textbf{AgentNetBench}}
    & \multicolumn{2}{c}{\textbf{AndroidControl}} \\
  \cmidrule(lr){2-3}\cmidrule(lr){4-5}
    & \textbf{20\%} & \textbf{40\%}
    & \textbf{20\%} & \textbf{40\%} \\
  \midrule
  Full KV                         & 25.1 & 25.1 & 55.4 & 55.4 \\
  \midrule
  Base                            & 14.1 & 17.2 & 41.2 & 45.8 \\
  \quad + MI                      & 15.4 & 18.1 & 47.4 & 46.2 \\
  \quad + TEM                     & 15.3 & 18.8 & 45.2 & 45.8 \\
  \quad + AEB                     & 16.0 & 18.9 & 44.0 & 46.8 \\
  \rowcolor{cyan!10}
  \textbf{STaR-KV (Ours)}         & \textbf{16.5} & \textbf{21.1} & \textbf{50.2} & \textbf{49.0} \\
  \bottomrule
  \end{tabular}
  }
\end{table}
\begin{table}[t]
  \centering
  \small
  \setlength{\tabcolsep}{4pt}
  \renewcommand{\arraystretch}{0.85}
  \caption{Ablation of temporal weighting functions in TEM. Cell
values are step accuracy (\%); column headers indicate the KV-cache
budget.}
  \label{tab:exp_ablation}
  \resizebox{\columnwidth}{!}{
  \begin{tabular}{@{}l cc cc@{}}
  \toprule
  \multirow{2}{*}{\textbf{Function}}
    & \multicolumn{2}{c}{\textbf{AgentNetBench}}
    & \multicolumn{2}{c}{\textbf{AndroidControl}} \\
  \cmidrule(lr){2-3}\cmidrule(lr){4-5}
    & \textbf{20\%} & \textbf{40\%}
    & \textbf{20\%} & \textbf{40\%} \\
  \midrule
  Full KV                        & 24.8 & 24.8 & 54.6 & 54.6 \\
  \midrule
  Gamma                          & 13.5 & 17.2 & 46.5 & 45.3 \\
  Linear                         & 14.8 & 19.5 & 48.1 & 47.0 \\
  \rowcolor{cyan!10}
  \textbf{Exponential (Ours)}    & \textbf{16.5} & \textbf{21.1} & \textbf{50.2} & \textbf{49.0} \\
  \bottomrule
  \end{tabular}
  }
  \vspace{-3mm}
\end{table}

\noindent\textbf{Ablation results.}
We further validate the key design choices of STaR-KV through
controlled ablations. All ablation experiments are conducted on
UI-TARS-1.5-7B~\cite{uitars}.

\noindent\textcircled{1} \textbf{Component ablation.}
Table~\ref{tab:main_ablation} shows that each calibration module
contributes a complementary gain over the Base configuration, which
uses the GQA-averaged attention score $\bar{s}_t$ with global
Top-$K$ selection and no calibration. On AgentNetBench, the three
modules progressively raise Base from $17.2\%$ to STaR-KV's $21.1\%$
at the $40\%$ budget (\textcolor{red}{$\uparrow$3.9}); a similar
pattern holds at $20\%$ ($14.1\!\to\!16.5\%$). On AndroidControl
under multi-frame context (\textbf{in contrast to the single-frame protocol
used in the main results Tab~\ref{tab:main_results}}), +MI alone delivers the largest individual
gain (\textcolor{red}{$\uparrow$6.2} at $20\%$), and the full model
reaches $50.2\%$. The three modules thus address complementary
bottlenecks. MI provides subspace-level spatial priors, TEM
suppresses stale multi-frame visual evidence, and AEB calibrates
score sharpness under varying attention reliability.

\noindent\textcircled{2} \textbf{Comparison of temporal decay functions.}
As shown in Table~\ref{tab:exp_ablation}, exponential decay is the most stable among the three temporal discount functions.
On AgentNetBench, it reaches $16.5\%$ at the $20\%$ budget (\textcolor{red}{$\uparrow$1.7} over linear, \textcolor{red}{$\uparrow$3.0} over gamma) and $21.1\%$ at $40\%$ (\textcolor{red}{$\uparrow$1.6} / \textcolor{red}{$\uparrow$3.9}).
On AndroidControl, it likewise ranks best in both settings, achieving $50.2\%$ at $20\%$ (\textcolor{red}{$\uparrow$2.1} / \textcolor{red}{$\uparrow$3.7}) and $49.0\%$ at $40\%$ (\textcolor{red}{$\uparrow$2.0} / \textcolor{red}{$\uparrow$3.7}).
This indicates that the \emph{shape} of temporal discounting matters for long-horizon GUI trajectories: linear decay penalizes history too aggressively and discards useful early interaction context, while gamma decay retains more long-range history but is consistently the weakest across both benchmarks.
Exponential decay strikes a smoother balance between suppressing stale visual evidence and preserving useful history, and is therefore adopted as the default TEM decay.

\noindent\textcircled{3} \textbf{Entropy-based vs.\ confidence-based AEB.}
As shown in Fig~\ref{fig:ablation_4panel} (a, b), entropy-based AEB
outperforms its confidence-based counterpart at every budget on
AgentNetBench, reaching $21.1\%$ vs.\ $19.5\%$ at $40\%$
(\textcolor{red}{$\uparrow$1.6}). The two variants share the same
KV budget and differ only in how the AEB temperature is estimated:
peak attention mass $p_{\max}$ versus normalized Shannon entropy of
the full distribution. This confirms that peak confidence is too local for long-horizon
trajectories with fluctuating attention, demanding a
distribution-level signal.

\noindent\textcircled{4} \textbf{Online vs.\ offline group prior.}
As shown in Fig~\ref{fig:ablation_4panel} (c, d), the online prior matches
the offline variant across budgets while removing its calibration
stage. On AgentNetBench, online is consistently stronger from $10\%$
onward, reaching $21.1\%$ vs.\ $19.4\%$ at $40\%$
(\textcolor{red}{$\uparrow$1.7}); on ScreenSpot-Pro, the two are
nearly identical. A fixed dataset-level prior is thus unnecessary:
online profiling estimates the prior during inference via an EMA of
per-group 2D MI, matching offline calibration at zero overhead.

\begin{figure}[t]
  \centering
  \includegraphics[width=\columnwidth]{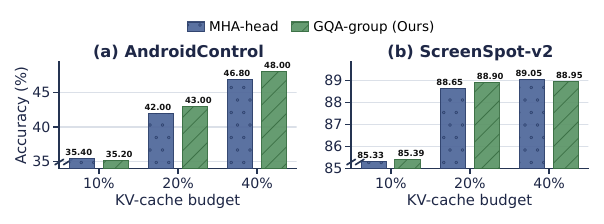}
  \caption{Ablation on subspace granularity for MI estimation under
UI-TARS-1.5-7B. \textbf{GQA-group} (default) estimates the prior at
the group level; \textbf{MHA-head} at the individual head level.
Cell values are accuracy (\%).}
  \label{fig:gqavsmha}
  \vspace{-3mm}
\end{figure}
\noindent\textcircled{5} \textbf{GQA-group vs.\ MHA-head granularity.}
Fig~\ref{fig:gqavsmha} compares estimating the MI prior at the
GQA-group level (default) versus the individual head level. To
simulate MHA-head on a GQA model, we replicate each shared key/value
across all query heads in the group~\cite{MHA} and estimate the MI
prior per head. GQA-group matches or outperforms MHA-head across
budgets (e.g., \textcolor{red}{$\uparrow$1.0} on AndroidControl at
$20\%$ and \textcolor{red}{$\uparrow$1.2} at $40\%$), as group-level
aggregation stabilizes the MI estimate by pooling statistics from
multiple heads. We adopt GQA-group for GQA models and per-head estimation for MHA, where the two coincide.

\subsection{Efficiency Analysis}

\begin{table}[t]
\centering
\small
\setlength{\tabcolsep}{4pt}
\renewcommand{\arraystretch}{0.85}
\caption{Decode-time efficiency on AgentNetBench with
UI-TARS-1.5-7B, measured as MFLOPs per generated token. STaR-KV
matches GUIKV within about $1\%$ across all settings.}
\label{tab:efficiency_decode_mflops}
\resizebox{\columnwidth}{!}{
\begin{tabular}{c l c c}
\toprule
\# Screenshots & KV Cache & $\gamma$ (\%) & MFLOPs / decoded token \\
\midrule
\multirow{5}{*}{3}
& Full Cache & 100 & 295.6 \\
\cmidrule(lr){2-4}
& GUIKV                   & 40  & 198.4 \textcolor{blue}{($-32.9\%$)} \\
\rowcolor{cyan!10}
\cellcolor{white} & \textbf{STaR-KV (Ours)}  & 40  & \textbf{196.9} \textcolor{blue}{($-33.4\%$)} \\
& GUIKV                   & 20  & 152.8 \textcolor{blue}{($-48.3\%$)} \\
\rowcolor{cyan!10}
\cellcolor{white} & \textbf{STaR-KV (Ours)}  & 20  & \textbf{153.5} \textcolor{blue}{($-48.1\%$)} \\
\midrule
\multirow{5}{*}{5}
& Full Cache & 100 & 378.7 \\
\cmidrule(lr){2-4}
& GUIKV                   & 40  & 240.5 \textcolor{blue}{($-36.5\%$)} \\
\rowcolor{cyan!10}
\cellcolor{white} & \textbf{STaR-KV (Ours)}  & 40  & \textbf{239.8} \textcolor{blue}{($-36.7\%$)} \\
& GUIKV                   & 20  & 186.2 \textcolor{blue}{($-50.8\%$)} \\
\rowcolor{cyan!10}
\cellcolor{white} & \textbf{STaR-KV (Ours)}  & 20  & \textbf{187.4} \textcolor{blue}{($-50.5\%$)} \\
\midrule
\multirow{5}{*}{10}
& Full Cache & 100 & 562.8 \\
\cmidrule(lr){2-4}
& GUIKV                   & 40  & 305.4 \textcolor{blue}{($-45.7\%$)} \\
\rowcolor{cyan!10}
\cellcolor{white} & \textbf{STaR-KV (Ours)}  & 40  & \textbf{307.1} \textcolor{blue}{($-45.4\%$)} \\
& GUIKV                   & 20  & 219.7 \textcolor{blue}{($-61.0\%$)} \\
\rowcolor{cyan!10}
\cellcolor{white} & \textbf{STaR-KV (Ours)}  & 20  & \textbf{218.5} \textcolor{blue}{($-61.2\%$)} \\
\bottomrule
\end{tabular}
}
\end{table}

\noindent\textcircled{1} \textbf{Decode-time FLOPs.}
We quantify efficiency by reporting decode MFLOPs per generated token
on UI-TARS-1.5-7B with AgentNetBench, varying the KV-cache budget
$\gamma$ and the number of screenshots
(Table~\ref{tab:efficiency_decode_mflops}). Compared with the full
cache, STaR-KV reduces decode MFLOPs by $33\%$ to $46\%$ at
$\gamma{=}40\%$ and $48\%$ to $61\%$ at $\gamma{=}20\%$, with savings
growing under image-heavy contexts. Despite introducing four
additional calibration modules, STaR-KV matches GUIKV in decode
efficiency to within about $1\%$ across all settings, as its
soft-global selection path replaces head-wise Top-$K$ with a single
token-level Top-$K$ and offsets the calibration overhead.

\begin{table}[t]
\centering
\small
\setlength{\tabcolsep}{5pt}
\renewcommand{\arraystretch}{0.95}
\caption{Peak GPU memory on ScreenSpot-Pro with UI-TARS-1.5-7B. STaR-KV
matches SnapKV and GUIKV within $\pm 0.05$\,GB at the highest
accuracy.}
\label{tab:memory}
\begin{tabular}{@{}l c c c@{}}
\toprule
\textbf{Method} & \textbf{Budget} & \textbf{Acc.\ (\%)} & \textbf{Peak Memory} \\
\midrule
Full Cache & 100\% & 41.68 & 37.36\,GB \\
\midrule
SnapKV   & 20\% & 39.22 & 22.99\,GB \\
SnapKV   & 40\% & 41.11 & 25.46\,GB \\
\midrule
GUIKV   & 20\% & 39.85 & 22.99\,GB \\
GUIKV   & 40\% & 40.73 & 25.46\,GB \\
\midrule
\rowcolor{cyan!10}
\textbf{STaR-KV (Ours)} & 20\% & \textbf{40.90} & \textbf{22.97}\,GB \\
\rowcolor{cyan!10}
\textbf{STaR-KV (Ours)} & 40\% & \textbf{41.70} & \textbf{25.43}\,GB \\
\bottomrule
\end{tabular}
\vspace{-2mm}
\end{table}
\noindent\textcircled{2} \textbf{Memory footprint.}
Table~\ref{tab:memory} reports peak GPU memory on ScreenSpot-Pro
with UI-TARS-1.5-7B. STaR-KV consumes essentially the same memory
as SnapKV and GUIKV at matched budgets (within $\pm 0.05$\,GB)
while achieving the highest accuracy, effectively alleviating cache
pressure under high-resolution multi-image inputs.

\section{Conclusion}
We presented \textbf{STaR-KV}, a training-free KV cache compression
framework that calibrates token importance along three axes:
subspace-aware scoring via online spatial mutual information, a
cumulative temporal stability discount that suppresses redundant
entries from persistently attended subspaces, and an entropy-derived
temperature that reshapes the score distribution. On four GUI agent benchmarks, STaR-KV outperforms state-of-the-art
KV compression methods (e.g., GUIKV, SnapKV) at matched budgets,
with essentially no compression-stage FLOPs ($-0.07\%$).

\section*{Limitations}
While STaR-KV achieves strong accuracy and memory trade-offs on GUI
agent benchmarks, several limitations remain.

\noindent\textbf{Empirical scope.}
Our evaluation focuses on two open-source GUI VLMs at the $7$B scale.
Validating STaR-KV on larger backbones and closed-source systems is
left for future work.

\noindent\textbf{Benchmark coverage.}
We evaluate on four established GUI-agent benchmarks. Extending the
evaluation to emerging benchmarks covering new domains or interaction
modalities is an interesting direction.

\noindent\textbf{Potential risks.}
As an efficiency-oriented technique, STaR-KV does not directly
introduce new safety hazards. However, by lowering the deployment
cost of GUI agents, it may indirectly facilitate the broader
deployment of agentic systems, including dual-use applications such
as automated web scraping or unauthorized UI automation. We
encourage users to apply STaR-KV in conjunction with existing
safeguards for responsible agent deployment.
\bibliography{custom}

\clearpage
\appendix
\section{Appendix}

\subsection{Implementation details}

This section summarizes the dataset-specific evaluation settings and STaR-KV hyper-parameters used in the main experiments. 
For all datasets, STaR-KV uses \texttt{rode\_group} with \texttt{soft\_global} selection and a recent window size of $W=8$. 
The online MI prior uses \texttt{mi\_saliency}, prior strength $\lambda=0.5$, MI-saliency mixture weight $\eta=0.5$, 5 online profiling steps, EMA decay $0.9$, temperature $\tau=1.0$, and a 10-step ramp for $\lambda_{\mathrm{eff}}$. 
All experiments use bfloat16 precision and FlashAttention-2.

\paragraph{Temporal settings.}
Across datasets, the temporal module uses exponential decay with $\delta=0.2$, $D_{\min}=0.1$, EMA decay $\rho=0.9$, and no warmup. 
For single-image benchmarks, this module reduces to the identity factor $D_t=1$.

\begin{center}
\scriptsize
\setlength{\tabcolsep}{3pt}
\renewcommand{\arraystretch}{0.95}
\captionof{table}{
Dataset-specific evaluation settings used in the main experiments.
}
\label{tab:dataset_eval_settings}
\begin{tabular}{lccc}
\toprule
Dataset & Input setting & \texttt{max\_new\_tokens} & Metric \\
\midrule
ScreenSpot-Pro & single image & 400 & accuracy \\
ScreenSpot-v2 & single image & 200 & accuracy \\
AndroidControl & single-step, single image & 600 & step accuracy \\
AgentNetBench & multi-step, image slots = 5 & 1000 & overall score \\
\bottomrule
\end{tabular}
\end{center}

\begin{center}
\scriptsize
\setlength{\tabcolsep}{3pt}
\renewcommand{\arraystretch}{0.95}
\captionof{table}{
Dataset- and model-specific STaR-KV hyper-parameters. 
AEB is applied at the layer level without temperature smoothing. 
All temporal settings use EMA decay $\rho=0.9$ and no warmup. 
For single-image benchmarks, the temporal module reduces to $D_t=1$.
}
\label{tab:dataset_hparams}
\begin{tabular}{llccc}
\toprule
Dataset & Model & AEB $T$ range & Temporal $\delta$ & Temporal $D_{\min}$ \\
\midrule
ScreenSpot-Pro & UI-TARS & $[0.75,1.25]$ & $0.2$ & $0.1$ \\
ScreenSpot-Pro & OpenCUA & $[0.95,1.05]$ & $0.2$ & $0.1$ \\
ScreenSpot-v2 & UI-TARS & $[0.75,1.25]$ & $0.2$ & $0.1$ \\
ScreenSpot-v2 & OpenCUA & $[0.95,1.05]$ & $0.2$ & $0.1$ \\
AndroidControl & UI-TARS & $[0.75,1.25]$ & $0.2$ & $0.1$ \\
AndroidControl & OpenCUA & $[0.95,1.05]$ & $0.2$ & $0.1$ \\
AgentNetBench & UI-TARS & $[0.95,1.05]$ & $0.2$ & $0.1$ \\
AgentNetBench & OpenCUA & $[0.95,1.05]$ & $0.2$ & $0.1$ \\
\bottomrule
\end{tabular}
\end{center}

\subsection{STaR-KV pipeline summary}
The overall procedure of STaR-KV is summarized in
Algorithm~\ref{alg:starkv}.
\begin{algorithm}[t]
\caption{STaR-KV: Spatio-Temporal Adaptive Re-weighting for KV Cache Compression in GUI Vision-Language Models}
\label{alg:starkv}
\small
\setlength{\abovedisplayskip}{2pt}
\setlength{\belowdisplayskip}{2pt}
\begin{algorithmic}[1]
\Require past KV $(K,V)$, query window $\{Q_{-i}\}_{i<W}$,
         recent KV $(K^{W}_{r},V^{W}_{r})$, visual span $\mathcal{V}$, budget $k$
\Ensure compressed cache $(K', V')$
\State $A \gets \mathrm{Pool}\!\big(\textstyle\sum_{i}\mathrm{softmax}(Q_{-i}K^{\!\top}\!/\sqrt{d})\big)$
\State $S[g,t] \gets \tfrac{1}{|\mathcal{H}_g|}\sum_{h\in\mathcal{H}_g} A[h,t]$
\State $\bar s_t \gets \tfrac{1}{G}\textstyle\sum_g S[g,t],\ \ g^{*}(t)\gets\arg\max_g S[g,t]$
\stage{stageMI}{Online MI prior}
\State {\color{stageMI}$\blacktriangleright$}\ $\mathrm{MI}_g \gets I(\mathrm{RankBin}(S[g,\mathcal{V}]);\,\mathrm{SpatialBin}(\mathcal{V}))$
\State {\color{stageMI}$\blacktriangleright$}\ $\overline{\mathrm{MI}}_g \gets \rho\,\overline{\mathrm{MI}}_g + (1-\rho)\,\mathrm{MI}_g$
\State {\color{stageMI}$\blacktriangleright$}\ $w_g \gets \mathrm{softmax}(\mathrm{minmax}(\overline{\mathrm{MI}}_g)/\tau)$
\State {\color{stageMI}$\blacktriangleright$}\ $s^{\mathrm{MI}}_t \gets \bar s_t\,[\,1-\lambda + \lambda\,G\,w_{g^{*}(t)}\,]$
\stage{stageTEM}{Temporal discount (skip if single-image)}
\State {\color{stageTEM}$\blacktriangleright$}\ $D_t \gets D_{\min}\!+\!(1{-}D_{\min})\,e^{-\delta\Delta_t\,\mathrm{stab}_{g^{*}(t)}}$
\State {\color{stageTEM}$\blacktriangleright$}\ $s^{\mathrm{tem}}_t \gets s^{\mathrm{MI}}_t\,D_t$
\stage{stageAEB}{Attention-Entropy Bridge}
\State {\color{stageAEB}$\blacktriangleright$}\ $p_t \gets \bar s_t / (\textstyle\sum_j \bar s_j + \epsilon)$
\State {\color{stageAEB}$\blacktriangleright$}\ $\hat H \gets -\textstyle\sum_t p_t\log(p_t{+}\epsilon)\,/\,\log L$
\State {\color{stageAEB}$\blacktriangleright$}\ $T \gets T_{\min}+(T_{\max}{-}T_{\min})\,\hat H$
\State {\color{stageAEB}$\blacktriangleright$}\ $s_t \gets \max(s^{\mathrm{tem}}_t,0)^{1/T}$
\stage{stageSEL}{Fixed-budget selection}
\State {\color{stageSEL}$\blacktriangleright$}\ $I \gets \mathrm{Sort}(\mathrm{TopK}(\{s_t\},k))$
\State {\color{stageSEL}$\blacktriangleright$}\ \Return $K'\!=\![K[I];K^{W}_{r}],\ V'\!=\![V[I];V^{W}_{r}]$
\end{algorithmic}
\end{algorithm}

\subsection{Detailed efficiency analysis}
We analyze efficiency at the KV-compression level, isolating it from
the full forward pass. STaR-KV reuses the standard attention
statistics of prior attention-based KV selection
methods~\citep{snapkv,h2o,pyramidkv}: recent-window queries attend
over past keys, followed by softmax, pooling, and score aggregation.
The dominant operations ($QK^{\top}$ and attention pooling) are
therefore shared with all such baselines.

STaR-KV further diverges from SnapKV in the selection stage. SnapKV
applies head-wise Top-$K$ over the attention cache, scaling as
$\mathcal{O}(HL\log k)$ with the number of heads $H$. STaR-KV instead
fuses head- and group-level signals into a single token-level score
vector and performs one \emph{soft-global} Top-$K$, reducing the
selection cost to $\mathcal{O}(L\log k)$. The saved head-axis factor
absorbs most of the calibration overhead introduced above.

Table~\ref{tab:flops} reports implementation-level FLOPs accounting
under the same cache budget. STaR-KV totals $155.29$M FLOPs against
SnapKV's $155.40$M, a net difference of $-0.07\%$. This indicates
that the added calibration is fully amortized by the cheaper
selection path.

This accounting reflects analytic FLOPs of the compression procedure,
not wall-clock latency, which additionally depends on kernel launch
overhead, memory access patterns, and sort/gather implementations.
The \emph{Selection} rows in Table~\ref{tab:flops} should likewise
be read as algorithmic complexity comparisons rather than iso-kernel
benchmarks. The takeaway is that STaR-KV improves token ranking
quality \emph{without} a meaningful FLOPs burden over attention-based
KV compression.

\begin{table}[t]
\centering
\small
\setlength{\tabcolsep}{6pt}
\renewcommand{\arraystretch}{1.1}
\caption{Implementation-level FLOPs accounting under the same KV-cache
budget. STaR-KV introduces four calibration modules
(GUI saliency, online MI profiling, temporal stability discount, AEB)
while \emph{reducing} selection cost from head-wise top-$k$ to a single
global top-$k$. The added overhead is fully offset by the cheaper
selection, yielding a net change of $-110{,}269$ FLOPs ($-0.07\%$).}
\label{tab:flops}
\resizebox{\columnwidth}{!}{%
\begin{tabular}{@{}l S[table-format=6.2] S[table-format=6.2] l@{}}
\toprule
\textbf{Component} & {\textbf{SnapKV}} & {\textbf{STaR-KV}} & \textbf{Note} \\
                   & {(MFLOPs)}        & {(MFLOPs)}         &                \\
\midrule
\multicolumn{4}{@{}l}{\textit{Shared backbone}} \\
\quad $QK^{\top}$                 & 143.36 & 143.36 & shared \\
\quad Softmax + Pool + Sum        &   2.38 &   2.38 & shared \\
\quad Gather + Concat             &   8.96 &   8.96 & shared \\
\midrule
\multicolumn{4}{@{}l}{\textit{STaR-KV calibration modules}} \\
\quad GUI saliency                & {--}   &   0.14 & added \\
\quad Online MI profiling         & {--}   &   0.17 & added \\
\quad Temporal stability discount & {--}   &   0.04 & added \\
\quad AEB sharpening              & {--}   &   0.02 & added \\
\midrule
\multicolumn{4}{@{}l}{\textit{Token selection}} \\
\quad Top-$k$ (head-wise vs. global) &   0.70 &   0.05 & different paths \\
\quad Fusion overhead                & {--}   &   0.17 & mean / group / scale / sort \\
\midrule
\textbf{Total} & \bfseries 155.40 & \bfseries 155.29 & \textcolor{darkred}{$-0.07\%$} \\
\bottomrule
\end{tabular}%
}
\end{table}

\subsection{Benchmark details}
\paragraph{ScreenSpot-Pro.}
ScreenSpot-Pro~\cite{screenspotpro} is evaluated as a single-image GUI grounding task. 
Each example contains one screenshot and one instruction, and the model predicts the target click or region. 
We use \texttt{max\_new\_tokens=400} and report overall grounding accuracy. 
Since only one visual span is provided, the temporal module degenerates to $D_t=1$. 
For UI-TARS-1.5-7B, AEB uses $T\in[0.75,1.25]$; for OpenCUA-7B, AEB uses $T\in[0.95,1.05]$.

\paragraph{ScreenSpot-v2.}
ScreenSpot-v2~\cite{screenspotpro} is also evaluated as a single-image grounding benchmark with all target categories. 
We use \texttt{max\_new\_tokens=200} and report overall accuracy. 
As in ScreenSpot-Pro, Temporal Stability Discount is inactive because the input contains only one visual span. 
The AEB ranges are shared with ScreenSpot-Pro: $[0.75,1.25]$ for UI-TARS-1.5-7B and $[0.95,1.05]$ for OpenCUA-7B.

\paragraph{AndroidControl.}
AndroidControl~\cite{androidcontrol} is evaluated as single-step action prediction under the high-level instruction setting. 
The main-table setting uses a single screenshot per step without historical frames. 
We use \texttt{max\_new\_tokens=600} and report step accuracy after action and argument normalization. 
Temporal discounting is therefore effectively inactive in the main-table setting. 
The AEB ranges are $[0.75,1.25]$ for UI-TARS-1.5-7B and $[0.95,1.05]$ for OpenCUA-7B.

\paragraph{AgentNetBench.}
AgentNetBench~\cite{opencua} evaluates multi-step GUI-agent trajectories. 
For UI-TARS-1.5-7B, we use multi-frame history with \texttt{image\_slots=5}; therefore, Temporal Stability Discount is active. 
STaR-KV states are reset at the trajectory level. 
We use \texttt{max\_new\_tokens=1000} and report the mean overall action score over the trajectory. 
For both UI-TARS-1.5-7B and OpenCUA-7B, AEB uses $T\in[0.95,1.05]$. 
For OpenCUA-7B, AgentNetBench is evaluated in a single-turn per-step setting to avoid degradation from accumulated dialogue history.
\subsection{Hyperparameter sensitivity.}
\begin{figure}[t]
  \centering
  \includegraphics[width=\columnwidth]{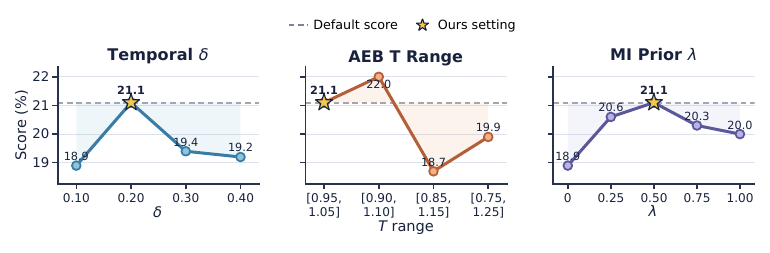}
  \caption{Hyperparameter sensitivity of STaR-KV on AgentNetBench
at the $40\%$ budget under UI-TARS-1.5-7B. We sweep the three main
hyperparameters: temporal decay $\delta$, AEB temperature range
$[T_{\min}, T_{\max}]$, and MI prior strength $\lambda$. The default
setting (marked by stars) sits at or near the peak of every curve,
with all configurations within $2.4$\,pp of the default.}
  \label{fig:hparam_sensitivity}
\end{figure}
Figure~\ref{fig:hparam_sensitivity} sweeps over the three main
hyperparameters of STaR-KV ($\delta$, $[T_{\min}, T_{\max}]$, and
$\lambda$) on AgentNetBench at the $40\%$ budget under
UI-TARS-1.5-7B. The default setting (marked by stars) sits at or
near the peak of every curve, and no configuration deviates from
the default by more than $2.4$\,pp, indicating that STaR-KV is
largely insensitive to its hyperparameter choices. The temporal
decay $\delta$ peaks at $0.2$, with both stronger ($0.1$) and weaker
($0.3$, $0.4$) decays leading to comparable drops, consistent with
its role as a balance between suppressing stale tokens and
preserving useful history. The AEB $T$ range $[0.95,1.05]$ is
competitive with the slightly wider $[0.90,1.10]$
(\textcolor{red}{$\uparrow$0.9}), while overly aggressive ranges
($[0.85,1.15]$, $[0.75,1.25]$) degrade slightly, suggesting that
moderate sharpening suffices once the MI and TEM modules have
already calibrated the score distribution. The MI prior strength
$\lambda$ exhibits the flattest sensitivity profile, peaking at
$\lambda=0.5$ with neighbouring values ($0.25$, $0.75$) within
$0.84$\,pp, confirming that the MI re-weighting operates as a soft,
scale-preserving prior rather than a dominant signal. These results
support adopting a single shared hyperparameter configuration
across all benchmarks.

\end{document}